\documentclass[lettersize,journal]{IEEEtran}
\usepackage{amsmath,amsfonts}
\usepackage{algorithmic}
\usepackage{algorithm}
\usepackage{array}
\usepackage[caption=false,font=normalsize,labelfont=sf,textfont=sf]{subfig}
\usepackage{textcomp}
\usepackage{stfloats}
\usepackage{url}
\usepackage{verbatim}
\usepackage{graphicx}
\usepackage{cite}
\usepackage{xcolor}
\newcommand{\dario}[1]{\textcolor{black}{#1}}
\newcommand{\hesam}[1]{\textcolor{black}{#1}}
\newcommand{\presub}[1]{\textcolor{black}{#1}}

\title{Brain-Body-Task Co-Adaptation can Improve \\Autonomous Learning and Speed \\of Bipedal Walking }

\author{Darío Urbina-Meléndez$^1$, Hesam Azadjou$^1$, Francisco J. Valero-Cuevas$^{*,1,2}$

\thanks{$^{1}$ D.U.-M, H.A. and F.V.-C are with the Alfred E. Mann Department of Biomedical Engineering, University of Southern California, Los Angeles, CA 90089 USA.
        {\tt\small [urbiname][azadjou][valero]@usc.edu}}\ 
\thanks{$^{2}$ F.V.-C is also with Division of Biokinesiology and Physical Therapy, University of Southern California, Los Angeles, CA 90089 USA.}
\thanks{$^{*}$ F.V.-C. is the corresponding author. [valero]@usc.edu}}

\begin{document}

\maketitle

\begin{abstract}

\dario{Inspired by} animals that co-adapt their brain and body to 
\dario{interact with the environment,} we present a \dario{tendon-driven and over-actuated (i.e., n joint, n+1 actuators)} bipedal robot that (i) exploits its backdrivable mechanical properties to manage \dario{body-environment interactions} without explicit control, \textit{and} (ii) uses \dario{a simple 3-layer neural network }to learn to walk after only 2 minutes of `natural' motor babbling \dario{(i.e., an exploration strategy that is compatible with leg and task dynamics; akin to childsplay).} 
This brain-body collaboration first learns to produce 
\presub{feet cyclical movements }`in air' and, without further tuning, can produce locomotion when \presub{the biped is} lowered to be in slight contact with the ground.
In contrast, training with 2 minutes of `na\"ive' motor babbling \dario{(i.e., an exploration strategy that ignores leg task dynamics)}, does not produce consistent 
\presub{cyclical movements} `in air', and produces erratic movements and no locomotion when in slight contact with the ground. 
When further lowering the biped 
\dario{and making the} desired leg trajectories reach 1cm below ground \dario{(causing the desired-vs-obtained trajectories error to be unavoidable)}, 
\presub{cyclical movements} based on either natural or na\"ive babbling presented almost equally persistent 
trends, and locomotion emerged with na\"ive babbling. 
Therefore, \dario{we show how continual learning of walking in unforeseen circumstances can be driven by continual physical adaptation rooted in the backdrivable properties of the plant and enhanced by exploration strategies that exploit plant dynamics.}
\dario{Our studies also demonstrate that the bio-inspired co-design and co-adaptations of limbs and control strategies can produce locomotion without explicit control of trajectory errors.} 

\end{abstract}

\section{Introduction}

Active and explicit control of robotic bipedal locomotion
poses multiple challenges, including\dario{: i)} hybrid dynamics that transition among single- and double-leg stances and aerial phases \cite{ames2018hybrid}, \dario{\textit{and} ii)} actuators with insufficient bandwidth to manage instantaneous impacts \cite{hurst2010actuator}. 
To address these challenges, studies that take inspiration from the musculature of organisms have incorporated mechanical components and architectures to \dario{reduce limb inertia by implementing cable (i.e., tendon) driven structures} \cite{urbina2021bio}, and increase the use of passive limb properties to manage impacts \cite{hurst2010actuator, rond2020mitigating}. 
Furthermore, approaches like Zero Moment Point (ZMP) enable balance during bipedal locomotion via quasi-static foot placements \cite{vukobratovic2004zero}, as in the ASIMO low-impact robot \cite{sakagami2002intelligent}\dario{, which is built and programmed in a way that avoiding impacts with the environment is one important design consideration.}
Truly agile robots need to break from this 'fear' of impacts and transition to more dynamical cases, theories like Hybrid Zero Dynamics have been developed where a reset map allows the system to go back to stable performance after the intrinsic impulse perturbations of ground interaction in dynamic behavior \cite{ames2018hybrid}. 
In the furthest extreme, there are robots whose own structure allows them to produce locomotion without feedback control (e.g. \cite{badri2022birdbot}). 
Proof of principle comes from passive walkers that can produce useful movements without sensors and/or actuators \cite{mcgeer1990passive, srinivasan2006computer}.

Largely missing from current approaches, however, is the most enviable capability of biological organisms: the ability to co-adapt their control strategies with their bodies to learn locomotion on their own.
Therefore, we focused on creating a \dario{tendon driven and over-actuated (i.e., 2 joints, 3 actuators) bipedal robot} that implements such brain-body co-adaptation to learn locomotion.
To do so, we combined two bio-inspired features: i) backdrivable limbs that adapt to \dario{environmental physical constraints}
(akin to musculotendons) \textit{and} ii) motor babbling compatible with leg and task dynamics, that allows brain-body collaboration through sparse physical actions (akin to childsplay \cite{fine2007trial,adolph2012you}), to heuristically learn to perform tasks \cite{yoon2018bayesian,kwiatkowski2019task,he2022convolutional}.

\dario{We present a } ``Natural'' motor babbling strategy as an extension of G2P or “General to Particular” model-agnostic algorithm \cite{marjaninejad2019autonomous} which enables bio-inspired learning of locomotion movements in tendon driven robotic limbs. 
This natural babbling strategy is an improvement of the na\"ive babbling strategy previously used by G2P.
\dario{Data collected during both, natural and na\"ive babbling, are used to train a simple 3 -layer artificial neural network (ANN) which represents the inverse map from 6D limb kinematics (i.e., for our robot proximal and distal joint position, velocities, and accelerations) to 3D motor control sequences (i.e., three motors actuating the joints through tendons).}

\dario{In \cite{marjaninejad2019autonomous}, it is seen how a na\"ive babbling strategy causes aproximately 80\% of the data generated to lie on edges of the configuration space, away from the area where the locomotion solutions lie. }
\dario{In contrast to this na\"ive strategy (that persistenly coactivates antagonist actuators, imposing movements that can conflict with leg dynamics), natural babbling resembles muscle mutual inhibition in living organisms \cite{day1984reciprocal,friesen1994reciprocal}.
This promotes a more informative sensory feedback, compatible with the limb properties.
In detail, when performing natural babbling, }
motor activations: 
i) produce joint rotations away from their limits of rotation and 
ii) follow a sinusoidal patterns instead of step functions (with a phase shift of 180 +-20 degrees for pairs of motors that act on the same joint, in other words two antagonist motors are not simultaneously activated with high activation values).
As a result the leg joints are more homogeneously exposed to the region of the configuration space where locomotion patterns lie, 
promoting a higher success rate of learning of robotic locomotion (ratio of experiments where walking is learned to those where it is not learned). 

We demonstrate with a physical robot that locomotion can emerge from the co-adaptation of actions learned from limited experience \hesam{(i.e. few shots of training)} enabled by backdrivable limbs that implicitly manage \dario{body-environment interactions.}
\hesam{The critical factors offered by this algorithm are the data efficiency and low-budget computational requirements, which can serve as a baseline for the lifelong learning of bipedal robots.}
Our study emulates the adaptive behavior of animals, where continual success of learned actions relies on useful brain-body-environment interaction \cite{chiel1997brain,valero2022bio}.

\section{Methods}
\subsection{Robot characteristics}
\label{methods:biped_characteristics}

\begin{figure}[!t]
\centering
\includegraphics[width=.85\linewidth]{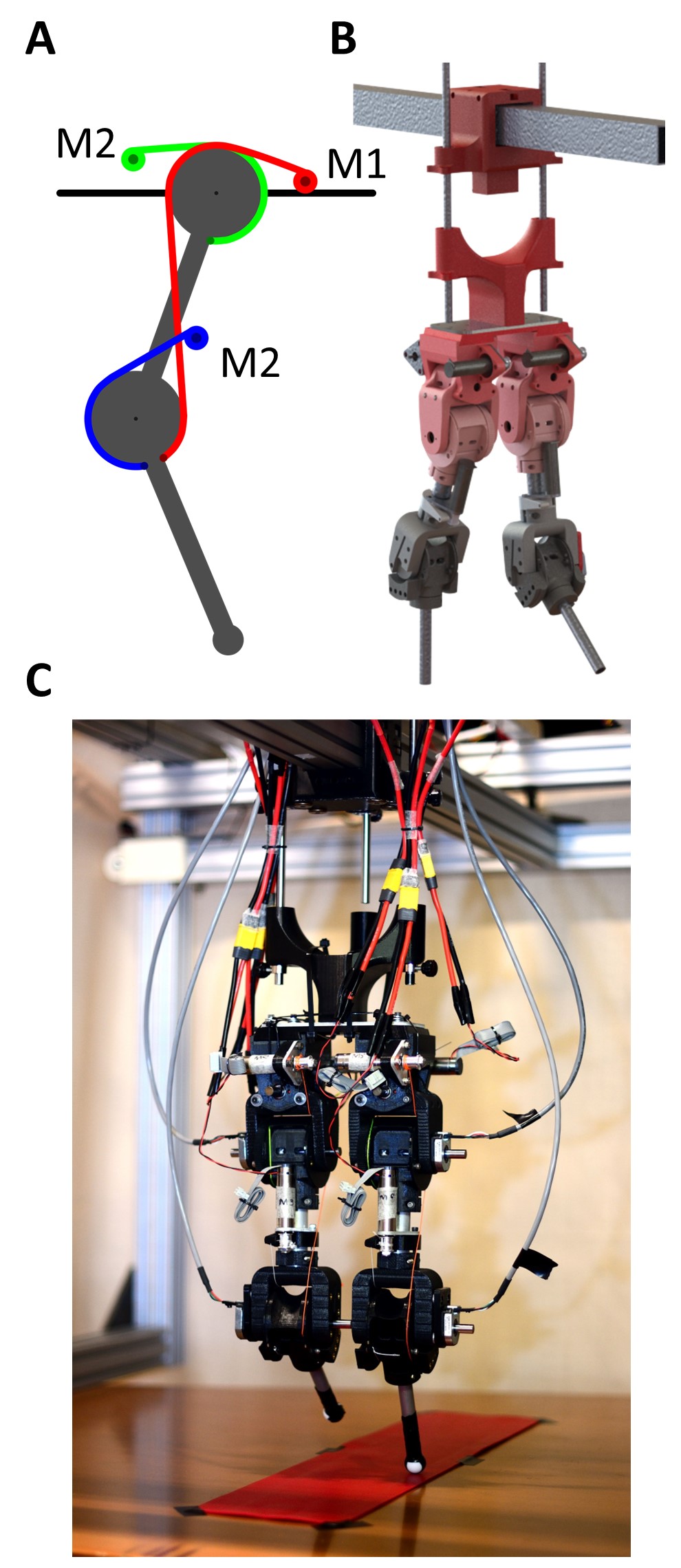}
\caption{\textbf{A-}Tendon route diagram of one leg, \textbf{B-} Render of the 3D model of the biped \textbf{C.-}  Photograph of the tendon-driven bipedal robot.
To reduce rotational inertia, motors M1 and M2 (Maxon DCX16S
GB KL 24V, 21:1 reduction ratio gearhead) are placed distally to the joints. 
}\label{fig:biped_hardware}
\end{figure}

For our experiments, we built and used a
tendon-driven physical bipedal robot (Figure \ref{fig:biped_hardware}). 
Each of its legs has hip and knee joints and a ball foot to facilitate the relative rotation of the lower section of the leg with respect to the ground. 

The mechanical power to the joints is provided by a structure that resembles a muscle: the force is provided by a motor, while the muscle-joint interface (which in our robot would be the motor-joint interface) is a string that we call tendon. 
This robot is over-actuated since it has more actuators than degrees of freedom (DoF). 
The tendon route is shown in Figure \ref{fig:biped_hardware}-A.

The tendon routing of our robot is an evolution of the routing for the robot in our already published paper \cite{marjaninejad2019autonomous}, where all the motors were placed distally to the leg (i.e., in the hip). Here we simplify the tendon routing by having only two motors placed distal to the leg  and one of them in the thigh. 
This design decision was made to reduce the torques driving the hip joint, thus potentially simplifying the task of learning a useful movement. 
The motors (Maxon DCX16S GB KL 24V) include a gearhead (with a reduction ratio of 21:1). 
Respectively, each  motor is called M1, M2, and M3, for details on their location please refer to Figure \ref{fig:biped_hardware}. Comparing two motors A and B, both set to the same voltage level and mechanical load; A with a gearhead and B without one: motor A reduces the back-drivability of the limb while increasing its mechanical power output capabilities. 
This is an advantage for when the design of the robot is changed to a heavier one due to a bigger body size and/or the addition of more components (e.g., sensors and actuators). 

The range of motion of the joints was bigger than for our previous robot designs, allowing us to explore the capability of the robot to track a desired trajectory independently of hard stops providing physical help. 
Here it is important to mention that in locomotion experiments the movement of a robot is typically physically limited by two components that serve as boundaries of its feasible configuration space: mechanical constraints (i.e., hard tops) in its own body, and environmental constraints (i.e, objects or ground itself). 
By designing our robot to have big ranges of motion normally not reachable while performing tasks 
(Figures \ref{fig:biped_desired_trajectories}-B and \ref{fig:babbling_data}-B), we focus on the role that environmental constraints have on the resultant performance of a task. 

To maintain rotational inertia as low as possible (having a direct impact on power consumption to meet the demands of leg movement), and to increase the stiffness of the legs, we used aluminum tubes as main components of the legs. 
We used additive manufacturing or 3D printing techniques, for the construction of the joints. 
We also considered the implementation of easy tendon attachment points to facilitate the replacement of tendons, which is the part of the robot that breaks more often.

We built a gantry to support the biped, only allowing its hip to move along the x and z axis in its sagittal plane. 
The gantry prevents the biped from falling down, allowing us to focus merely on the task of learning a locomotion cycle.

\subsection{General G2P overview}
\label{methods:naive_natural}

\begin{figure*}
\centering
\includegraphics[width=6.2in]{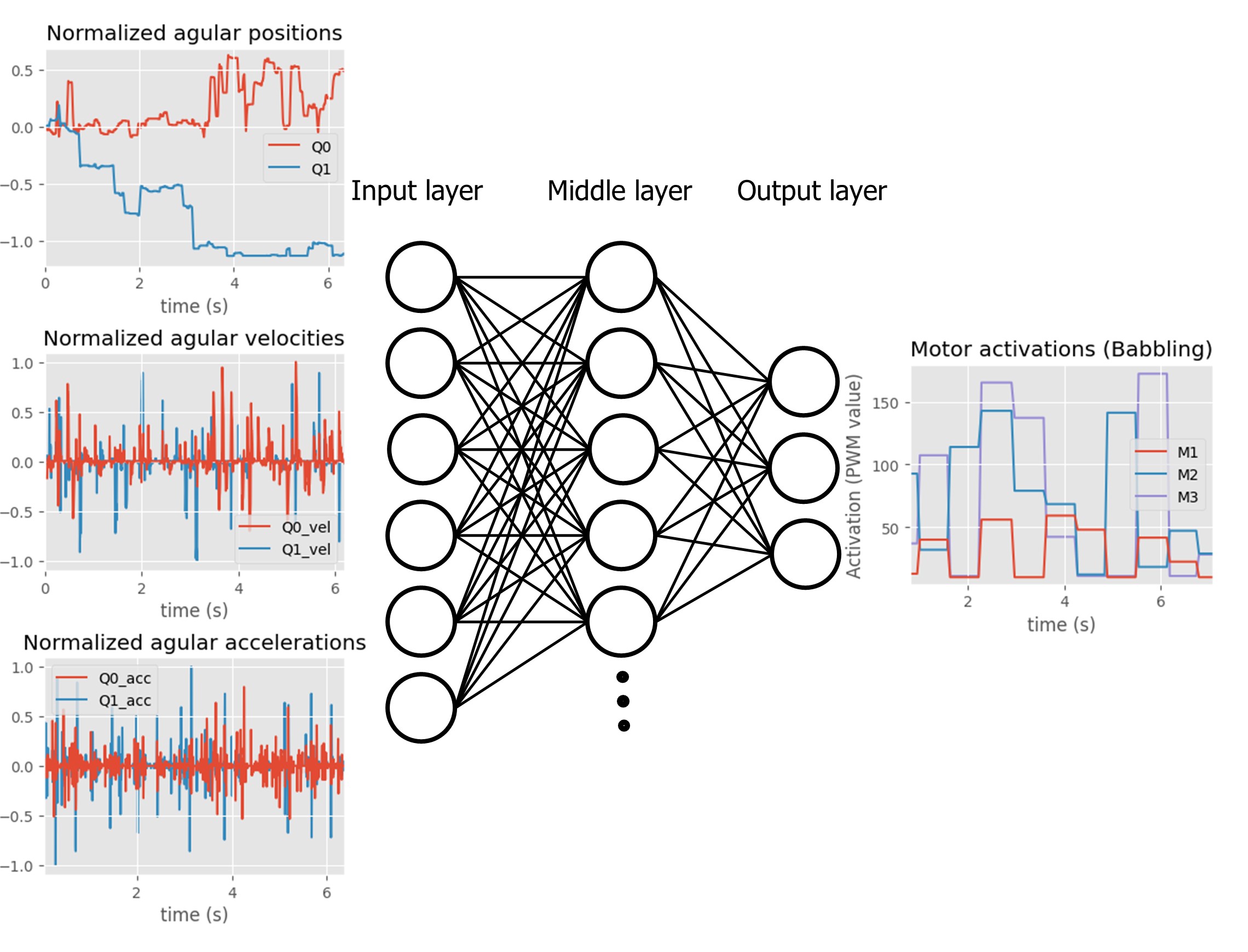}
\caption{
Representation of the ANN used as a map from six limb kinematics (input nodes, left column) to 3 motor activations (output nodes, right column). 
In this figure we show real data used to train the ANN (particularly na\"ive babbling data)
As a reminder, motor activations in babbling are random (Specific details on na\"ive babbling are given in sections:\ref{methods:naive_natural}).
The ANN has three fully connected layers: input, hidden and output layers with respectively 6, 15 and 3 nodes.
Note that motors are persistently simultaneously activated  (i.e., coactivation, see motors M2 and M3 in leftmost panel), this decreases the spread in training data. 
}
\label{fig:neural_network_naive}
\end{figure*}

The first version of the learning algorithm that we use was developed in \cite{marjaninejad2019autonomous}, it is called the General to Particular (G2P) algorithm.
This algorithm uses an Artificial Neural Network (ANN) as a map from inputs to outputs (respectively desired kinematics to motor activations) \dario{(Figures \ref{fig:neural_network_naive} and \ref{fig:neural_network_natural})}. 
The ANN is trained with input-output data sets obtained from babbling and tested 
with input-output data sets obtained from babbling by predicting outputs given inputs. 
The predicted outputs are compared with ground truth motor activations outputs. 
The difference between predicted and obtained values is the error and the goal is to reduce such error. 
The testing/training data set size ratio is 0.25. 
We use one ANN per leg.
In \cite{marjaninejad2019autonomous}, G2P refines this map with a reinforcement learning approach, for this paper we do not consider such a section of the algorithm since we are interested in understanding the value of the data obtained during babbling.

The ANN used for Natural and Na\"ive G2P (Figures \ref{fig:neural_network_naive} and \ref{fig:neural_network_natural} respectively) represents the inverse map from 6D limb kinematics (i.e., for our robot proximal and distal joint position, velocities, and accelerations) to 3D motor control sequences (i.e., three motors actuating the joints through tendons), it has three fully connected layers (input, hidden and output layers) with 6, 15 and 3 nodes, respectively. 

As the transfer functions for all nodes, we selected the hyperbolic tangent sigmoid function, which is an S-like function that produces a bounded output value in a range between -1 and 1. Additionally, we chose this function over the sigmoid since the gradient of the second is 
bigger than the first. 
The higher gradient produces a greater sensitivity to changes in the input values, producing higher updates in the weights of the networks (thus potentially faster learning).
We also applied a scaling for the output layer (giving values between -1--1) to obtain values to cover the whole motor control range values (0-255).

The weights and biases were initialized based on the Nguyen–Widrow initialization algorithm \cite{nguyen1990improving,wayahdi2019initialization}; with this, we avoid initializing weights close to the regions where the gradient of the transfer function has very small or high values. 
Having initial values localized in the mentioned region creates undesired output saturation. 
To obtain the best results, this approach randomly initializes weights close to the midpoint of the transfer function (i.e., 0 for the cases of our experiments). 

As a performance/error function, we used the mean square error (m.s.e.) approach. 
With this, the mean of the differences between values predicted by the ANN and the ground truth values are calculated. \hesam{This loss function aims to minimize the overall prediction error.}

This error is propagated backward to update the initial weights, the action performed with the Levenberg–Marquardt back propagation technique, the assignment of new weights is particularly done with Adaptive Moment Estimation (Adam), a gradient descent method chosen over MomeNtum, AdaGrad, RMSProp. 
Adam is the standard go-to method since it includes benefits from both Momentum and RMSProp. 
To find the best model weights, it leverages the usually seen speed of MomeNtum, and adaptability to gradients with different orientations commonly well handled by RMSProp. 
Each time the backpropagation is complete, it is considered that an epoch happened. 
We determined the maximum number of epochs to 100; also, the model training stops after there is no improvement after 5 epochs.

\subsection{Natural babbling: changes to G2P babbling strategy}
\label{methods:natural}

As mentioned in the introduction, we made changes to the babbling strategy of G2P to more homogeneously expose the leg joints to the areas in its configuration space where locomotion patterns lie. 
To keep our focus on assessing the usefulness of the data to produce a mapping with which a desired trajectory can be tracked, we particularly tested the G2P capability to create motor activations to limb kinematics map without any refinement to such a map.
With this paper, we show that (for a two DoF, three actuators leg) properly obtained data can be enough to train an ANN to produce useful movement (more details in results and discussion sections).

\begin{figure*}
\centering
\includegraphics[width=6.8in]{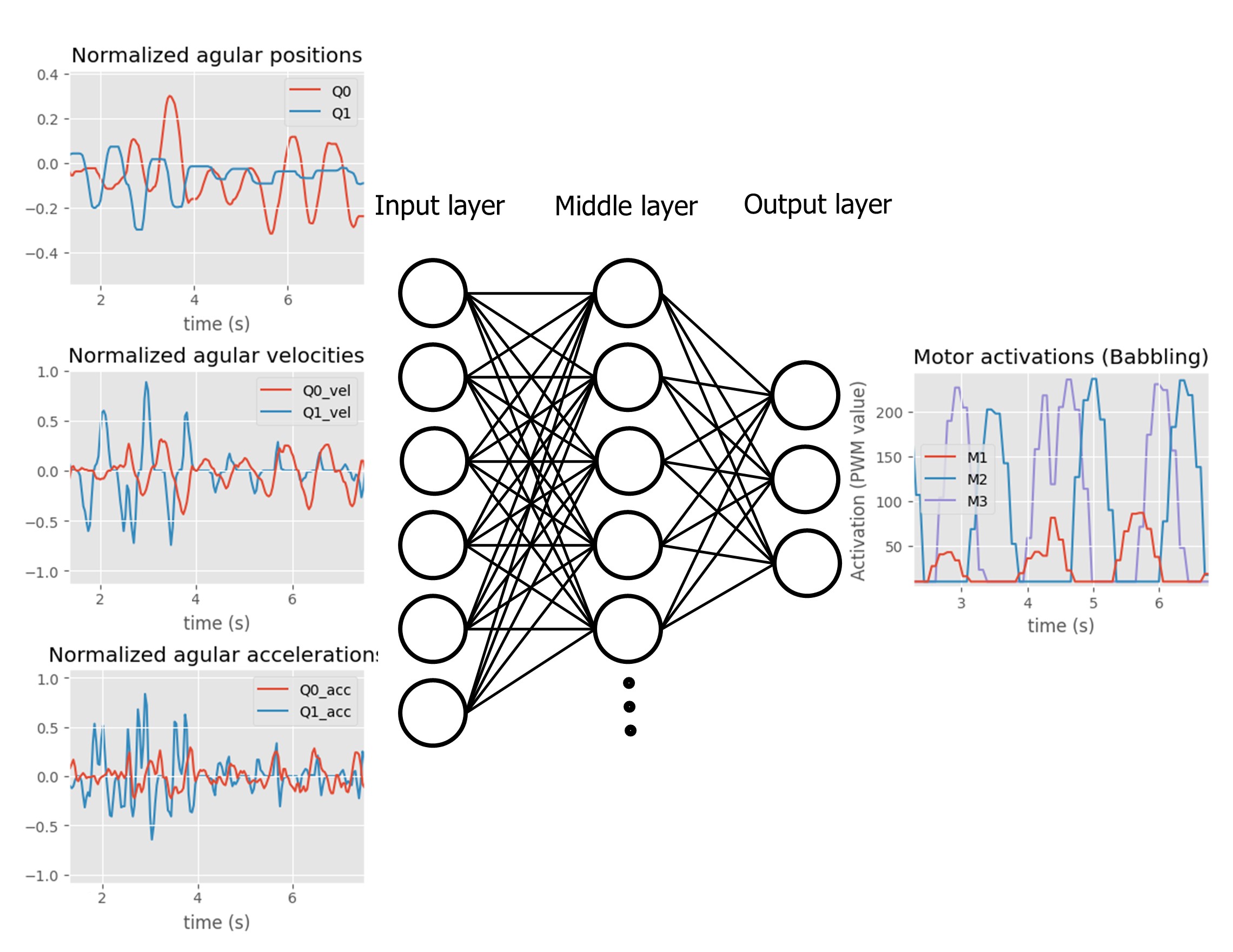}
\caption{
Representation of the ANN used as a map from six limb kinematics (input nodes, left column) to 3 motor activations (output nodes, right column). 
In this figure we show real data used to train the ANN (particularly natural babbling data)
As a reminder, motor activations in babbling are random (Specific details on natural babbling are given in section:\ref{methods:natural}).
The ANN has three fully connected layers: input, hidden and output layers with respectively 6, 15 and 3 nodes.
This figure shows how oscillatory movements are produced (see "Normalized angular positions" panel) driven by the by oscillatory babbling activations (rightmost panel) with significant difference activation level between antagonist motors.
}
\label{fig:neural_network_natural}
\end{figure*}

\dario{Before explaining the details of natural babbling, it is important to highlight that}
na\"ive babbling consists of random step PWM signal variation for each one of the motors \dario{and that } 
each motor signal is independent of the others (frequency of steps change:  ~1.3 Hz), \dario{as shown in Figure \ref{fig:neural_network_naive}, rightmost panel.} 
\dario{For natural babbling, w}e modified 
the randomness of motor activations by including the rule that the activation level of two antagonist motors should be significantly different \dario{(As observed in motors (M) 2 and 3 in Figure \ref{fig:neural_network_natural}, rightmost panel)}. 

For natural babbling (Figure \ref{fig:neural_network_natural}), each PWM signal for each of the motors follows a sinusoid profile. 
Considering that the mean value of the signals is 0, only the positive section is used. 
For each motor, the signal amplitude is varied randomly. 
M1 and M2 signals have a phase shift of 180 deg.
This is to avoid simultaneous activations of the motors which cause no hip movement to happen [\cite{day1984reciprocal,friesen1994reciprocal}]. 
Every 15 seconds, the phase between M1 and M3 was increased by 36 deg and the baseline of each signal varies +-30 PWM units (approximately +- 1V). To get a sinusoid-like shape, 
steps in series need to be considered (this is a digital system, so we are discretizing the signal). 
Step frequency: ~6 Hz. Sinusoid frequency (every time a period is completed):  ~.6 Hz. 
Frequency of each signal peak: ~1.3 Hz. 
Each peak (natural babbling) has approximately the same width as each step of na\"ive babbling. 
All frequencies are reported as approximate values. 
It is intrinsic to the microcontroller behavior to have slight variations in signaling and sampling frequency. 
The limits of rotation of each of the joints were never reached with natural babbling, a crucial point for our results and conclusions (Figure \ref{fig:biped_desired_trajectories}-B)

\subsection{Desired foot trajectory characteristics and variations}

\label{methods:foot_trajecotry_and_biped_support}

\begin{figure}
\centering
\includegraphics[width=3.4in]{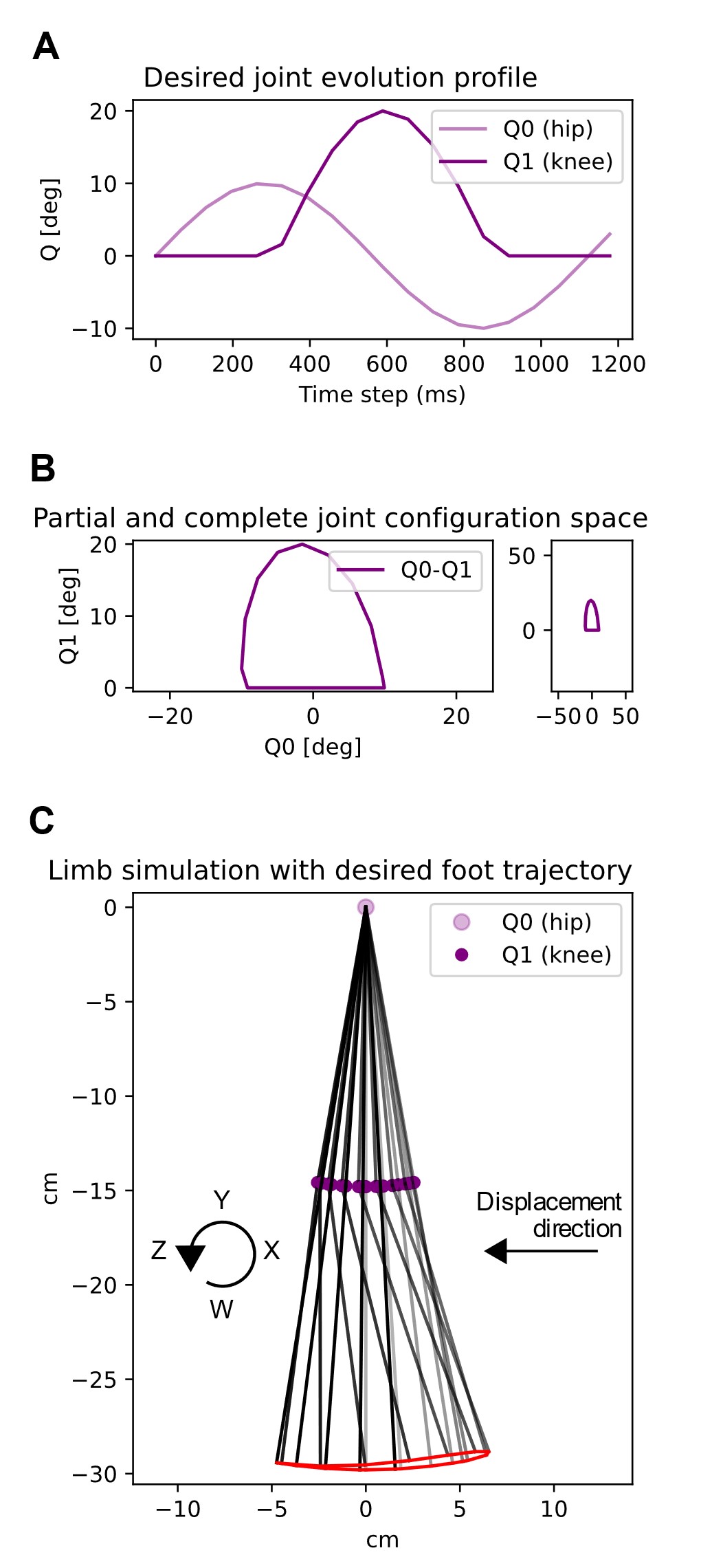}
\caption{Desired joint and foot trajectories. 
\dario{
In \textbf{A} it is shown a hip and knee joint evolution profile that produces limb movements away from the limits of its in-air configuration space as shown in \textbf{B} (Both panels in B represent the same, right one is a zoom out version). The resultant foot trajectory, shown in \textbf{C} is such that permits its front and back swings to have different height (necessary point to produce locomotion).
Note that the desired foot trajectory is always kept at a constant distance from the hip; thus if the hip position is changed the desired foot trajectory will also change.  
}}
\label{fig:biped_desired_trajectories}
\end{figure}

Before hardware experiments were performed, we did a forward kinematics analysis of possible 
\dario{limb movements that allowed us to obtain the desired joint evolution profile and ranges shown in Figure \ref{fig:biped_desired_trajectories}-A and B).} 
\dario{The resultant} 
foot trajectory is such that allows its front and back swings to have different heights (Figure \ref{fig:biped_desired_trajectories}-C).

As shown in Figure~\ref{fig:in_air_on_grond}, we divide our experiment into three main conditions 
determined by the location of the desired trajectory with respect to the ground.
\dario{The desired trajectory always has the same distance to the robot's hip, changing the position of the desired trajectory requires changing the robot's hip height by re-configuring the gantry that prevents the robot from falling down (Figure~\ref{fig:biped_hardware}-C).}
\dario{Depending on its location, a fraction or no part of the desired trajectory is reachable by the feet of the biped. We divide our experiments in three cases:}

\begin{enumerate}
\item\textbf{Condition 1: Desired trajectories in air-} 
only in air movement, with no 
\dario{interaction} with the ground. 
When performing movements, the feet trajectories will be limited only by the characteristics of the biped itself (Figure ~\ref{fig:in_air_on_grond}-A). 

\item \textbf{Condition 2: Desired trajectories in slight contact with the ground- }
desired foot trajectories are \dario{only} partially reachable since they are partially under the ground level. 
In other words, ground constraints the movement of the robot to stay over the boundary marked by the ground
(Figure~\ref{fig:in_air_on_grond}-B). 

\item \textbf{Condition 3: Desired trajectories 1 cm under the ground
-}
desired trajectories are  unreachable, they are completely under the ground level.
This is the condition where the biped’s movements are more constrained. Also, for this condition, the area of the feasible joint configuration space is smaller than in points 1 or 2 (i.e. here the biped movements are constrained to exist between the limits imposed by the ground and the limits marked by the limits of joint rotations) 
(Figure ~\ref{fig:in_air_on_grond}-C).\end{enumerate}

\subsection{Hardware experiments steps}

The following steps were performed using both: naïve and natural babbling. 
Eight trials of this experiment were performed, four based on na\"ive babbling and four on natural babbling.
If the biped displaces its body mass for 40 cm we consider this a successful walking trial.
The success rate is 
\dario{calculated by dividing the number of successful trials by the number of performed trials of a particular kind (i.e., condition and type of babbling data used).}
When a result is reported as “mean”, it is the average value from four trials. 
For the mean cases of spread and detrended fluctuation analysis, the number of values considered is eight (left and right legs for each of four trials: total eight). 

\begin{figure*}[!t]
\centering
\begin{center}
\includegraphics[width=1\linewidth]
{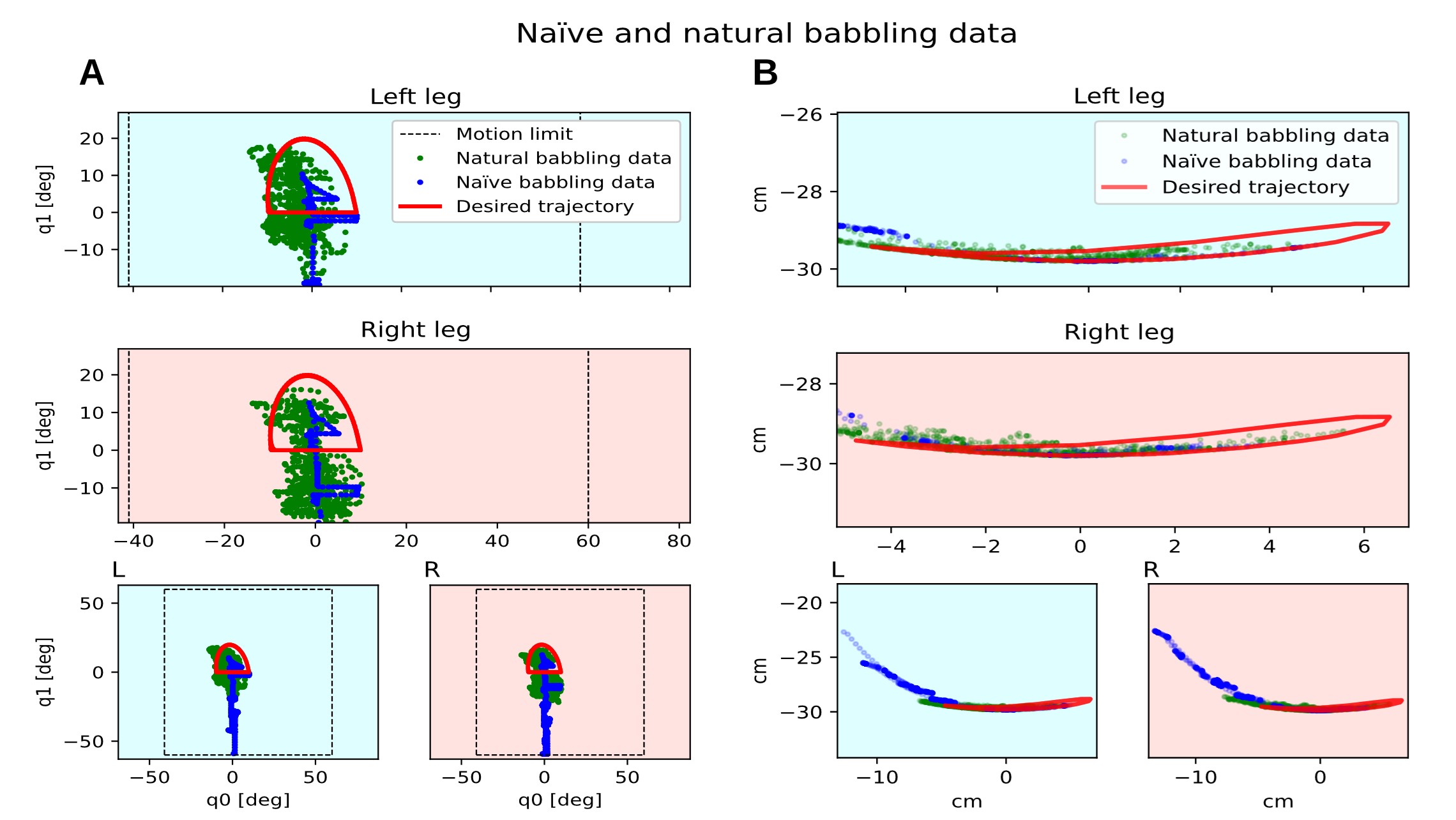}
\end{center}
\caption{Two minutes of babbling data and desired trajectories for one trial. \textbf{A}: Joint Space with joint motion limits marked with a doted square, \textbf{B}: Endpoint Space. 
The spread values are 0.14 and 0.18
for na\"ive babbling data (left and right legs respectively) and 0.60 and 0.53 for natural babbling data (left and right legs respectively)
}
\label{fig:babbling_data}
\end{figure*}

These are the steps we followed to perform our experiments:

\begin{enumerate}
\item Collect babbling data for two minutes (Figure \ref{fig:babbling_data}). Babbling characteristics are described in  Section \ref{methods:naive_natural}. 
\item Train an ANN to map motor activations to limb kinematics as described in Section \ref{methods:naive_natural}. 
\item With desired trajectories in air (i.e., Section \ref{methods:foot_trajecotry_and_biped_support}, biped suspended in air, no ground constraint), track the desired foot trajectory (Figure ~\ref{fig:in_air_on_grond}-A). 
\item  With desired trajectories in slight contact with the ground (i.e., Section \ref{methods:foot_trajecotry_and_biped_support}, biped's hip at 40 cm off the ground). Perform trajectory tracking as in (Figure~\ref{fig:in_air_on_grond}-B). 
Measure the time the biped takes to travel 40 cm in case there is successful walking. 
\item With desired trajectories 1 cm under the ground (i.e., Section \ref{methods:foot_trajecotry_and_biped_support}, biped's hip at 39 cm off the ground, Figure ~\ref{fig:in_air_on_grond}-C). 
Measure the time the biped takes to travel 40 cm in case there is successful walking. 

\end{enumerate}

\iftrue
\subsection{Data analysis (Spread calculation)}

We discretized the area within the desired trajectory into $1\times1 mm^{2}$ pixels and checked if the foot visited that pixel during a single babbling trial. 
Then by calculating the ratio of the occupied pixels to all pixels, we quantified the spread. Spread quantifies how well the algorithm (specifically, the babbling) can explore different kinematics by knowing the locations that feet have passed through.

\subsection{Data analysis (Detrended Fluctuation Analysis)}
\label{DFA_analisis}
In Detrended Fluctuation Analysis (DFA), the fractal scaling component estimates a time series' scaling behavior which represents the power law scaling behavior of the time series over various time scales. The steps for DFA are as follows:
\begin{enumerate}
  
 \item First, we detrended the time series data of the endpoint's 
 distance to the hip from each trial
 by dividing the time series into non-overlapping windows of equal length and then fitted a polynomial function of first degree to each window.
 \item Then we divided the detrended series into smaller segments of equal length (boxes). The scale factor determines the length of the boxes.
 \item Afterward, we calculated the root-mean-square fluctuation (F) for each box in the detrended series.
 \item Then, we calculated the average root-mean-square fluctuations across all the boxes at a given scale.
 \item We repeated steps 1 to 4 for different scale factor values and plotted the average fluctuation versus the scale factor (DFA curve). 
 \item Finally, we analyzed the DFA curve to check the time series data for long-term correlations. The DFA curve shows a power-law relationship between the fluctuation and the scale factor quantified by the slope alpha (fractal scaling component) using linear regression on a log-log scale.
\end{enumerate}

A higher fractal scaling component indicates that the time series exhibits stronger long-term correlations or persistence over various time scales, which means that the fluctuations in the time series at larger time scales are more correlated, and the time series has a more persistent trend.
Conversely, for a lower fractal scaling component, this analysis indicates weaker long-term correlations or anti-persistence in the time series, which means that the fluctuations at larger time scales are less correlated, and the time series has a less persistent trend \cite{peng1995quantification,peng1994mosaic,ihlen2012introduction}. We use the persistence of trends and strength of correlation in the legs' movements as a criterion to compare how well and robustly the biped walks (in case walking is achieved) in different cases and conditions.
\fi

\section{Results}

\subsection{Exploiting limb mechanical properties \dario{increases} the spread of training data and increases success rate of locomotion learning} 

All results reported in this subsection correspond to babbling data and walking attempts 
for Condition 2: Desired trajectories in slight contact with the ground
(Figure~\ref{fig:in_air_on_grond}-B).
\dario{As a reminder, the success rate is calculated by dividing the number of successful trials by the number of performed trials of a particular kind (i.e., condition and type of babbling data used).}

Two minutes of natural babbling data are enough to produce locomotion, while 2 minutes of na\"ive babbling data are not enough (Figure~\ref{fig:in_air_on_grond}-B). 
With natural babbling 
G2P learned walking in 75\% of the trials compared to 0\% for na\"ive babbling trials. 
Mean displacement speed for successful natural-babbling-based trials was 1.9 cm/sec. Speed for 3 out of 4 successful trials: 2.45, 1.96, 1.3 cm/sec. 

The difference, as previously described, between the na\"ive and natural cases resides in the babbling data. 
More spread babbling data (i.e., natural babbling data are more spread compared to na\"ive babbling data, as shown in Figure \ref{fig:babbling_data}) shows that the babbling was more successful in exploring the leg kinematics, which is the primary purpose of babbling. 
Consequently, compared to natural cases, a lower success rate 
happen when training with na\"ive babbling data.

As shown in Figure \ref{fig:babbling_data}, natural babbling data are closer to the regions of the configuration space where locomotion solutions lie. 
If we analyze the spread of this data within the area delimited by a desired trajectory, we see that the spread for the natural babbling data is higher than that of the na\"ive babbling data. 
For the trial presented in Figure \ref{fig:babbling_data}, left-right leg spread of na\"ive babbling data: 0.14 and 0.18 respectively; left-right leg 
spread of natural babbling data: 0.60 and 0.53 respectively. Mean spread values for na\"ive and natural babbling data respectively are: 0.55 and 0.95. 

In \cite{aggarwal2001surprising} it is described how a model to be able to describe a system, and to accurately predict its behavior, needs to be trained with more training samples spanning throughout the entire range of possible values such samples could possibly have.
In our experiments 
most of the na\"ive babbling points lie away and few inside the desired trajectory,  
in many cases failing on training a model that can accurately predict the behavior inside the desired trajectory. 
In this work behavior will be the motor commands to pull on the tendons to produce cyclical movements that are close to the desired trajectory. 
This is seen in Figure \ref{fig:in_air_on_grond}-A where the blue trajectories based on a model trained with na\"ive babbling data fail to closely resemble the desired trajectory. 
In contrast natural babbling points, which \dario{are more spread and} lie inside of the desired trajectory are better to train a model which can predict the motor activations required to produce cyclical foot trajectory patterns that better resemble the desired trajectory. 
This is seen in Figure \ref{fig:in_air_on_grond}-A where the green trajectories based on a model trained with natural babbling data better resemble the desired trajectory compared to the case of the na\"ive babbling based experiments.

\subsection{Placing desired trajectories completely under ground level increases walking success rate and produces faster walking}

\begin{figure*}[!t]
\centering
\includegraphics[width=1\linewidth]
{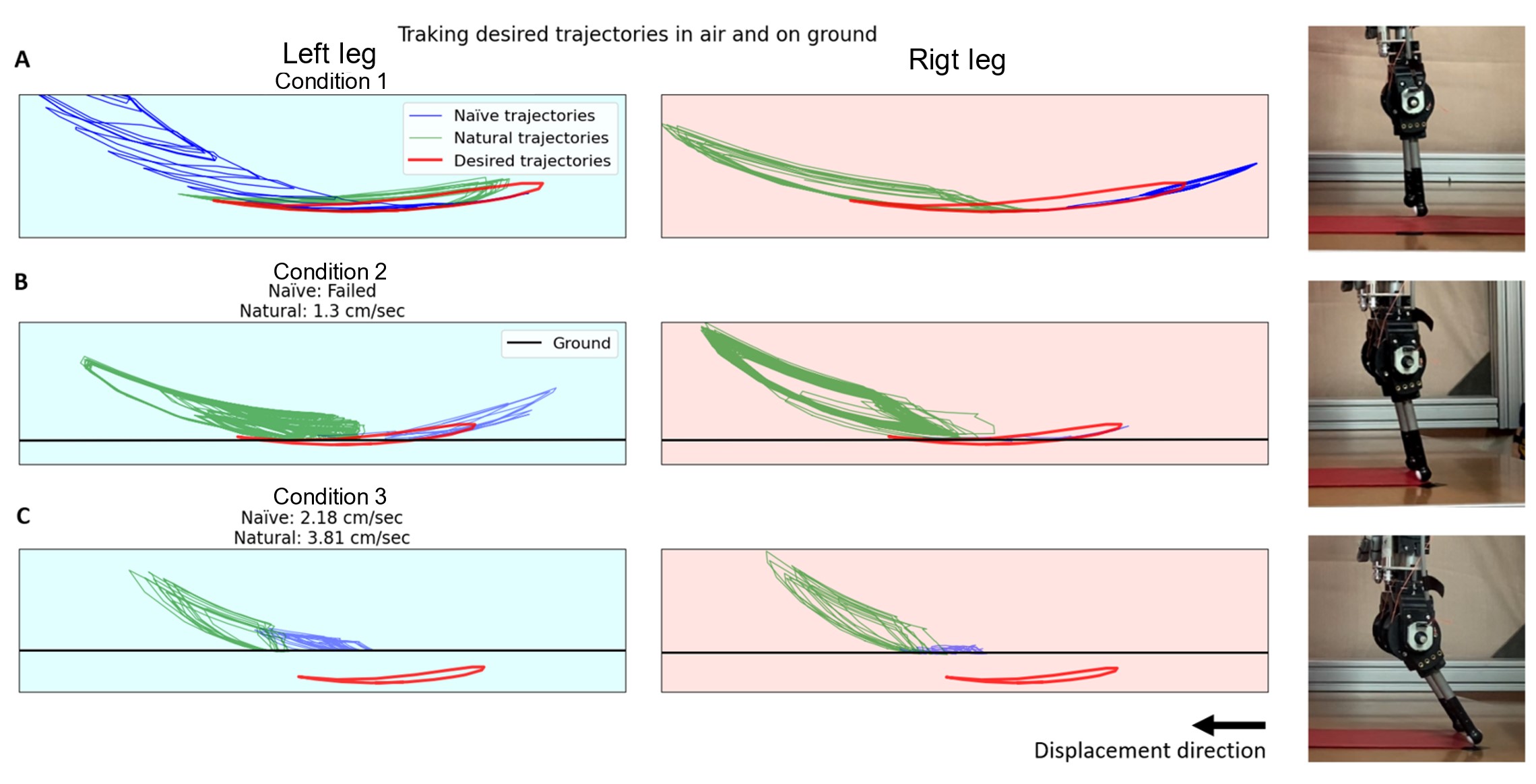}
\caption{Plots of obtained and desired foot trajectories shown together with close ups of the biped feet in different conditions. ``Na\"ive and Natural trajectories'' means trajectories obtained from training with na\"ive and natural babbling (this figure corresponds to the same trial as the one presented in Figure\ref{fig:babbling_data}): \textbf{Condition 1 (A)}: Desired trajecotries in air, \textbf{Condition 2 (B)}: Desired trajectories in slight contact with the
ground (locomotion emerges when training with natural but not with na\"ive babbling) and \textbf{Condition 3 (C)}: Desired trajectories 1cm under ground (locomotion emerges when training with both natural and na\"ive babbling). 
}
\label{fig:in_air_on_grond}
\end{figure*}

When the desired trajectories were 1 cm under the ground (Condition 3) (Figure~\ref{fig:in_air_on_grond}-B), G2P learned supported bipedal walking in 100\% of the trials based on both na\"ive and natural babbling.
Na\"ive case speeds (1.79, 3.27, 1.7, 2.18 cm/sec), natural case speeds (5.03, 4.93, 6.19, 3.81 cm/sec)
Respectively, mean displacement speeds for this cases were 2.23 cm/sec and 4.99 cm/sec. 
For trials based on natural babbling, when going from the condition where the desired trajectories are in slight contact with the ground \dario{(Condition 2)} to the condition where the desired trajectories are 1 cm under ground,
mean speed increased by 262\%, and success rate was increased from 75\% to 100\%. 
For the trials based on naive babbling the success rate was increased from 0 to 100\%. 

For the condition where the desired trajectories are in slight contact with the ground \dario{(Condition 2),}
the biped can only barely touch ground with fully straight legs, reducing the work that the legs produce to only the swing of the hip. 
In contrast, when the desired trajectories are 1 cm under ground \dario{(Condition 3),} the biped 
can produce work with both hip swing and knee flexion (Figure \ref{fig:in_air_on_grond}). 

Compared to the in-air performance of the biped (desired trajectories in air
), when the desired trajectories are in slight contact with the ground,
the scaling behavior (See Detrended Fluctuation Analysis in Methods, Section \ref{DFA_analisis}) for all our experiments drops (Figure \ref{fig:DFA_FIGURE}). 
This shows, as expected, that following the trend on the ground for the biped is more complicated than in the in-air condition. 
When trained with na\"ive babbling data and when
the desired trajectories are 1 cm under ground (compared to when the desired trajectories are slightly in touch with the ground)
, the generated movement shows significantly higher scaling components (p approximately of 0.03)
, indicating more persistent locomotion.
On the other hand, 
when trained with natural babbling data and when
the desired trajectories are 1 cm under ground (compared to when the desired trajectories are slightly in touch with the ground)
, there is not a significantly different scaling component (p approximately of 0.22); however, there is less variance from trial to trial.

For cases trained with natural babbling data, when taking the desired trajectory from slightly touching ground to being completely under ground, there is an increment in walking speed. 
The reason for this is that, for cases based on natural babbling training data, walking has already emerged when the desired trajectory slightly touches ground
. In the other hand, for cases based on na\"ive babbling training data, walking first emerges 
when the desired trajectories are placed 1 cm under ground.
Both na\"ive and natural cases present an improvement when 
trajectories are placed under ground level,
but na\"ive cases 
has 
less improvement after locomotion emerges than cases based on natural babbling. 

\section{Discussion}

This paper aims to motivate the creation of  bipedal robots that learn locomotion via data-driven 
co-adaptation with the dynamics of the plant 
\dario{to manage interactions with the environment.}
This is made possible by using motor babbling to inform a motion planning strategy that produces 
\presub{cyclical movements} that can undergo useful adaptations thanks to the backdrivable and impact-resilient properties of the legs. 
These properties allow the unsupervised modification of a previously learned behavior to enable the emergence of locomotion under different (previously unseen) conditions. 
We find that a bio-inspired approach to `natural' motor babbling compatible with the dynamics of the tendon-driven legs improves the success of locomotion learning and performance compared to `na\"ive' arbitrary motor babbling.
The techniques presented here could be further complemented by other relevant approaches such as the calculation of parameters useful to maintain a balanced gait such as  
zero moment point (ZMP)  \cite{vukobratovic2004zero}, or  hybrid zero dynamics (HZD) \cite{ames2018hybrid} that explicitly considers the transitions between locomotor contact states. 
Even though these techniques are not necessary for the successful performance of our robot, in general they are potential options to further complement the experiments of this paper which do not focus on balance, but particularly on the generation of useful 
\presub{cyclical movements} 
for locomotion. 

A central aspect of our results is that the robot's backdrivable limbs
\dario{interact with the environment by allowing their}
movements to adapt to where the desired trajectory of a walking action is located with respect to the ground: in air, in partial contact with the ground (partially reachable) or under ground level (unreachable). 
For each of these conditions, interference 
\dario{of the desired trajectory} with the ground 
was 
progressively greater,
and the adaptation of a previously learned action was automatically modulated. 
Thus, the success of the resulting behavior does not depend on explicitly modulating or reducing errors. 
Rather---similar to the adaptive behavior observed in the locomotion of insects \cite{full1999templates}, crustaceans \cite{herreid1986locomotion}, and birds \cite{daley2007proxdist}---successful locomotion emerges because of, and not in spite of, brain (or controller)-body-environment interactions \cite{chiel1997brain}. 
This adaptation happened with a performance strategy not explicitly aware of interference or impacts with the environment.

\begin{figure*}[!t]
\centering
\includegraphics[width=1\linewidth]
{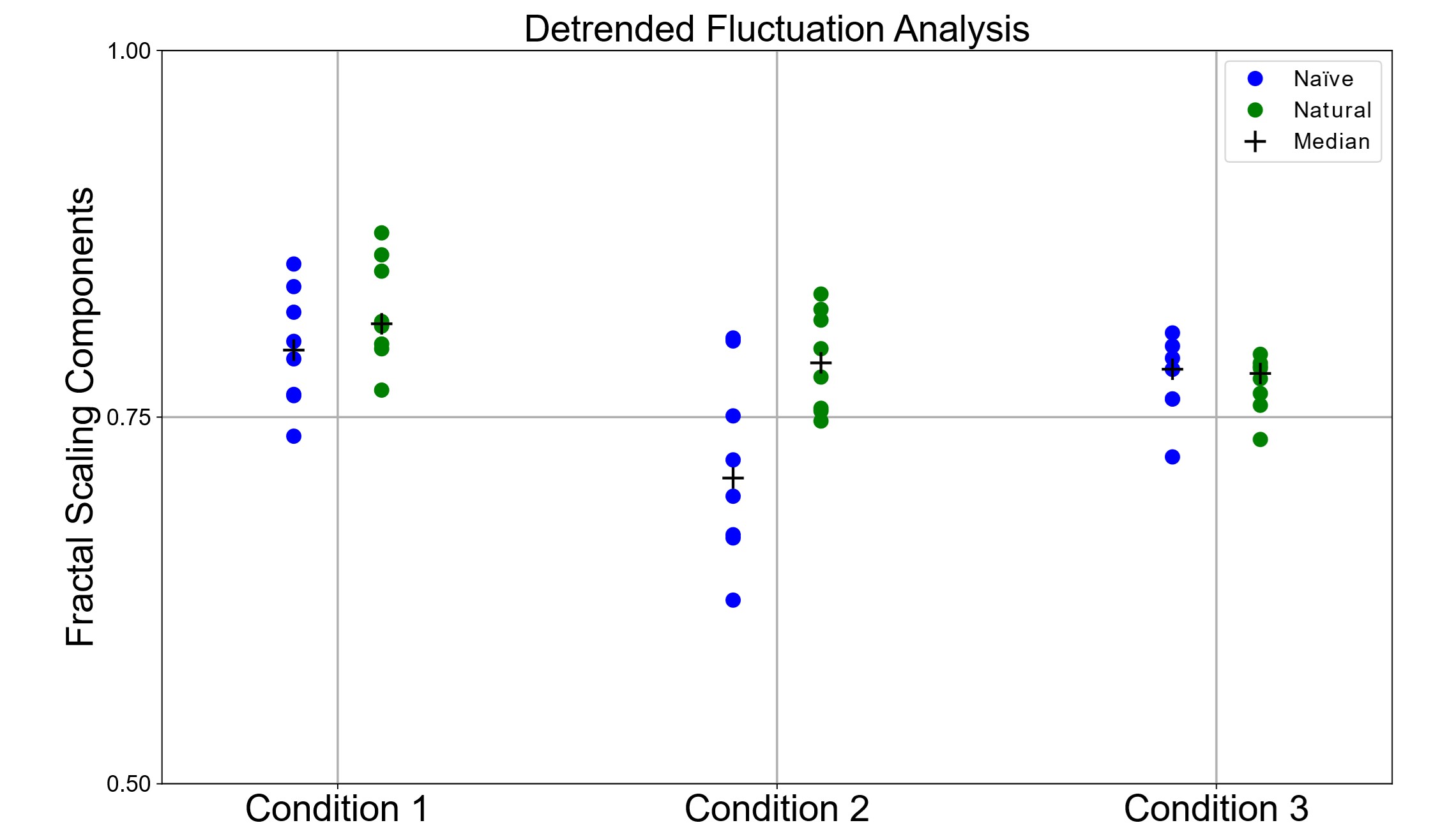}
\caption{This figure shows the dot plot of the fractal scaling components for na\"ive (blue) and natural (green) cases from eight different trials (four trials from each leg). Conditions 1, 2, and 3  show fractal scaling components when the desired trajectory is in the air, in slight contact with the ground, and 1 cm under the ground, respectively.
}
\label{fig:DFA_FIGURE}
\end{figure*}

For the natural babbling case (compared to na\"ive case), we found higher fractal scaling components for 
\presub{cyclical movements} with ground interference, as shown in Figure~\ref{fig:DFA_FIGURE}-Condition 2, suggesting they are more persistent.
In the case of na\"ive babbling, increased ground interference had a more profound effect. When there was slight contact with the ground, we saw no locomotion and lower fractal scaling components (Figure~\ref{fig:DFA_FIGURE}-Condition 2). But when contact with the ground was further increased (Figure~\ref{fig:DFA_FIGURE}-Condition 3), locomotion emerged from 
\presub{cyclical movements} with higher fractal scaling components comparable to those for natural babbling.
These results points to counterintuitive controller-body-environment interactions that produce better locomotion as interference with the ground increases. While we expected the reduction of the workspace of the leg to hamper locomotion, it seems the compliance of the legs (due to their backdriveability) adapt sufficiently well to shape the limits cycles to produce locomotion without the control signals being explicitly aware of 
control trajectory errors.

Another fundamental aspect of this study is that we prescribed a type of motor babbling (i.e., natural motor babbling) that is compatible with, and exploits the bio-inspired mechanical properties of the tendon-driven limbs.
Although similar in principle to Berniker et al. \cite{berniker2009simplified}, where the anatomical properties of a bio-inspired limb are exploited, we develop these strategies directly in hardware (and not in simulation). Moreover, we do not explicitly simplify the task by prescribing recurring muscle patterns (i.e., muscle synergies) to produce limb movements. It is our natural motor babbling that implicitly finds useful patterns of motor activations to the tendons. 
In fact, our natural motor babbling is one of the important  extensions to out prior work on autonomous learning of locomotion \cite{marjaninejad2019autonomous}. By using this type of motor babbling that tends to avoid  antagonist motor commands, we take inspiration from biological organisms where co-contraction can be energetically wasteful.
Actions that leverage the backdrivable mechanical properties of the plant, compatible with the over-and under-determined actuation of its tendon-driven limbs, are parallel to one of the fundamental blocks of limb function \cite{valero2016fundamentals,valero2022bio} to produce oscillatory limb movements (e.g., leg swin \cite{friesen1994reciprocal}) .

\section{Conclusion}
\dario{We made changes to the training babbling strategy of G2P to more homogeneously expose a biped’s leg joints to the areas in its configuration space where locomotion patterns lie. 
We did that by implementing a natural babbling strategy that exploits the tendon-driven bio-inspired mechanical properties of its limbs (i.e. oscillatory movements produced by oscillatory activations, with significant difference activation level between antagonist motors). 
We observed that natural babbling reduces the spread of training data and increases the success rate of locomotion learning when environmental constraints are minimal (Condition 2 of our experiments). 
Furthermore we also observed that increasing environmental constraint to the system (interference between ground and desired trajectories) increased the tendency of the plant to behave homogeneously between different trials (regardless of trials being based on natural or na\"ive babbling). 
This shows how, even though the environment (i.e., ground) generates a higher desired vs. obtained trajectory errors, it also collaborates with the backdrivable biped legs by  “guiding” them to perform a successful task by reducing their feasible configuration space.}  

We present proof-of-principle that effective locomotion can emerge from brain-body-environment interactions
\dario{driven by a controller that does not aim to reduce errors with respect to desired locomotion trajectories}. 
We find that these effective interactions arise from the co-adaptation facilitated by bio-inspired backdrivable properties of limbs. 
Moreover, the 
\presub{cyclical movements} motor commands are informed by pseudo-random motor babbling that exploits and leverages the bio-inspired tendon-driven mechanical and dynamical properties of the limbs.
This demonstrates the bio-inspired co-design and co-adaptations of limbs and control strategies can produce locomotion without explicit control of trajectory errors.

\section{Supplementary material}
$https://github.com/DarioUrbina/natural\_babbling$

\section*{Conflict of Interest Statement}

The authors declare that the research was conducted in the absence of any commercial or financial relationships that could be construed as a potential conflict of interest.

\section*{Author Contributions}

DU-M lead: the writing of this manuscript, conceptualization of the studies, building of the hardware for the experiments, experimentation, result analysis, figure making and discussion. DU-M performed experiments and did data collection. HA performed the Detrended Fluctuation Analysis and its figure, participated in the discussions to analyse results and helped writing parts of the methods and introduction of the manuscript. FV-C adviced DU-M on the initial conceptualization of the studies and provided general direction for the project. All persons designated as authors qualify for authorship, and all those who qualify for authorship are listed.

\section*{Funding}
Support for DU-M was provided by the joint research fellowship granted by Consejo Nacional de Ciencia y Tecnolog\'ia and the Viterbi School of Engineering (CONACYT-Mexico) at the University of Southern California (USC).
Support to HA was provided by the Graduate School of USC through the Provost Fellowship. 
Research reported in this publication was supported in part by the National Institute of Neurological Disorders and Stroke of the National Institutes of Health under award number R21NS11361, Department of Defense CDMRP Grant MR150091, DARPA-L2M program grants W911NF1820264 and W911NF2120070, and National Science Foundation Collaborative Research in Computational Neuroscience under award number CRCNS Japan-US 2113096 to FJV-C. 
The content is solely the responsibility of the authors and does not necessarily represent the official views of
the National Institutes of Health, the National Science Foundation, the Department of Defense, or DARPA.

\section*{Acknowledgments}
The authors \dario{would like to} thank Daniel Wang, Irie Cooper and Yifan Xue for their support in designing and manufacturing the physical system and electronics. 
Author DU-M would like to thank professor James Finley for discussions on the paper rationale as well as for providing detailed feedback on the methods and relevance of the findings.
Author DU-M would like to thank professor Nicolas Schweighofer for pointing out the weak points and opportunities of improvement of the presented studies.
The authors \dario{would also like to} thank Suraj Chakravarthi Raja for providing valuable technical and scientific insights about the experiments reported in this publication.
The authors \dario{would like to} thank Grace Niyo for her support with proofreading the paper manuscript.
The authors acknowledge the access to equipment for building experimental hardware provided by the Baum Family Makerspace, from the Viterbi School of Engineering.


\newpage

\bibliographystyle{IEEEtran}
\bibliography{paper.bib}

\begin{thebibliography}{10}
\providecommand{\url}[1]{#1}
\csname url@samestyle\endcsname
\providecommand{\newblock}{\relax}
\providecommand{\bibinfo}[2]{#2}
\providecommand{\BIBentrySTDinterwordspacing}{\spaceskip=0pt\relax}
\providecommand{\BIBentryALTinterwordstretchfactor}{4}
\providecommand{\BIBentryALTinterwordspacing}{\spaceskip=\fontdimen2\font plus
\BIBentryALTinterwordstretchfactor\fontdimen3\font minus \fontdimen4\font\relax}
\providecommand{\BIBforeignlanguage}[2]{{%
\expandafter\ifx\csname l@#1\endcsname\relax
\typeout{** WARNING: IEEEtran.bst: No hyphenation pattern has been}%
\typeout{** loaded for the language `#1'. Using the pattern for}%
\typeout{** the default language instead.}%
\else
\language=\csname l@#1\endcsname
\fi
#2}}
\providecommand{\BIBdecl}{\relax}
\BIBdecl

\bibitem{ames2018hybrid}
A.~D. Ames and I.~Poulakakis, ``Hybrid zero dynamics control of legged robots,'' 2018.

\bibitem{hurst2010actuator}
J.~W. Hurst, J.~E. Chestnutt, and A.~A. Rizzi, ``The actuator with mechanically adjustable series compliance,'' \emph{IEEE Transactions on Robotics}, vol.~26, no.~4, pp. 597--606, 2010.

\bibitem{urbina2021bio}
D.~Urbina-Mel{\'e}ndez, D.~Wang, and F.~Valero-Cuevas, ``Bio-inspired tendon-driven robotic limbs,'' \emph{(No Title)}, p.~60, 2021.

\bibitem{rond2020mitigating}
J.~J. Rond, M.~C. Cardani, M.~I. Campbell, and J.~W. Hurst, ``Mitigating peak impact forces by customizing the passive foot dynamics of legged robots,'' \emph{Journal of Mechanisms and Robotics}, vol.~12, no.~5, p. 051010, 2020.

\bibitem{vukobratovic2004zero}
M.~Vukobratovi{\'c} and B.~Borovac, ``hybrid zero-moment point—thirty five years of its life,'' \emph{International journal of humanoid robotics}, vol.~1, no.~01, pp. 157--173, 2004.

\bibitem{sakagami2002intelligent}
Y.~Sakagami, R.~Watanabe, C.~Aoyama, S.~Matsunaga, N.~Higaki, and K.~Fujimura, ``The intelligent asimo: System overview and integration,'' in \emph{IEEE/RSJ international conference on intelligent robots and systems}, vol.~3.\hskip 1em plus 0.5em minus 0.4em\relax IEEE, 2002, pp. 2478--2483.

\bibitem{badri2022birdbot}
A.~Badri-Spr{\"o}witz, A.~Aghamaleki~Sarvestani, M.~Sitti, and M.~A. Daley, ``Birdbot achieves energy-efficient gait with minimal control using avian-inspired leg clutching,'' \emph{Science Robotics}, vol.~7, no.~64, p. eabg4055, 2022.

\bibitem{mcgeer1990passive}
T.~McGeer \emph{et~al.}, ``Passive dynamic walking,'' \emph{Int. J. Robotics Res.}, vol.~9, no.~2, pp. 62--82, 1990.

\bibitem{srinivasan2006computer}
M.~Srinivasan and A.~Ruina, ``Computer optimization of a minimal biped model discovers walking and running,'' \emph{Nature}, vol. 439, no. 7072, pp. 72--75, 2006.

\bibitem{fine2007trial}
M.~S. Fine and K.~A. Thoroughman, ``Trial-by-trial transformation of error into sensorimotor adaptation changes with environmental dynamics,'' \emph{Journal of neurophysiology}, vol.~98, no.~3, pp. 1392--1404, 2007.

\bibitem{adolph2012you}
K.~E. Adolph, W.~G. Cole, M.~Komati, J.~S. Garciaguirre, D.~Badaly, J.~M. Lingeman, G.~L. Chan, and R.~B. Sotsky, ``How do you learn to walk? thousands of steps and dozens of falls per day,'' \emph{Psychological science}, vol.~23, no.~11, pp. 1387--1394, 2012.

\bibitem{yoon2018bayesian}
J.~Yoon, T.~Kim, O.~Dia, S.~Kim, Y.~Bengio, and S.~Ahn, ``Bayesian model-agnostic meta-learning,'' \emph{Advances in neural information processing systems}, vol.~31, 2018.

\bibitem{kwiatkowski2019task}
R.~Kwiatkowski and H.~Lipson, ``Task-agnostic self-modeling machines,'' \emph{Science Robotics}, vol.~4, no.~26, p. eaau9354, 2019.

\bibitem{he2022convolutional}
Y.~He, C.~Zang, P.~Zeng, Q.~Dong, D.~Liu, and Y.~Liu, ``Convolutional shrinkage neural networks based model-agnostic meta-learning for few-shot learning,'' \emph{Neural Processing Letters}, pp. 1--14, 2022.

\bibitem{marjaninejad2019autonomous}
A.~Marjaninejad, D.~Urbina-Mel{\'e}ndez, B.~A. Cohn, and F.~J. Valero-Cuevas, ``Autonomous functional movements in a tendon-driven limb via limited experience,'' \emph{Nature machine intelligence}, vol.~1, no.~3, pp. 144--154, 2019.

\bibitem{day1984reciprocal}
B.~Day, C.~Marsden, J.~Obeso, and J.~Rothwell, ``Reciprocal inhibition between the muscles of the human forearm.'' \emph{The Journal of physiology}, vol. 349, no.~1, pp. 519--534, 1984.

\bibitem{friesen1994reciprocal}
W.~O. Friesen, ``Reciprocal inhibition: a mechanism underlying oscillatory animal movements,'' \emph{Neuroscience \& Biobehavioral Reviews}, vol.~18, no.~4, pp. 547--553, 1994.

\bibitem{chiel1997brain}
H.~J. Chiel and R.~D. Beer, ``The brain has a body: adaptive behavior emerges from interactions of nervous system, body and environment,'' \emph{Trends in neurosciences}, vol.~20, no.~12, pp. 553--557, 1997.

\bibitem{valero2022bio}
F.~J. Valero-Cuevas and A.~Erwin, ``Bio-robots step towards brain--body co-adaptation,'' \emph{Nature Machine Intelligence}, vol.~4, no.~9, pp. 737--738, 2022.

\bibitem{nguyen1990improving}
D.~Nguyen and B.~Widrow, ``Improving the learning speed of 2-layer neural networks by choosing initial values of the adaptive weights,'' in \emph{1990 IJCNN international joint conference on neural networks}.\hskip 1em plus 0.5em minus 0.4em\relax IEEE, 1990, pp. 21--26.

\bibitem{wayahdi2019initialization}
M.~Wayahdi, M.~Zarlis, and P.~Putra, ``Initialization of the nguyen-widrow and kohonen algorithm on the backpropagation method in the classifying process of temperature data in medan,'' in \emph{Journal of Physics: Conference Series}, vol. 1235, no.~1.\hskip 1em plus 0.5em minus 0.4em\relax IOP Publishing, 2019, p. 012031.

\bibitem{peng1995quantification}
C.-K. Peng, S.~Havlin, H.~E. Stanley, and A.~L. Goldberger, ``Quantification of scaling exponents and crossover phenomena in nonstationary heartbeat time series,'' \emph{Chaos: an interdisciplinary journal of nonlinear science}, vol.~5, no.~1, pp. 82--87, 1995.

\bibitem{peng1994mosaic}
C.-K. Peng, S.~V. Buldyrev, S.~Havlin, M.~Simons, H.~E. Stanley, and A.~L. Goldberger, ``Mosaic organization of dna nucleotides,'' \emph{Physical review e}, vol.~49, no.~2, p. 1685, 1994.

\bibitem{ihlen2012introduction}
E.~A. Ihlen, ``Introduction to multifractal detrended fluctuation analysis in matlab,'' \emph{Frontiers in physiology}, vol.~3, p. 141, 2012.

\bibitem{aggarwal2001surprising}
C.~C. Aggarwal, A.~Hinneburg, and D.~A. Keim, ``On the surprising behavior of distance metrics in high dimensional space,'' in \emph{Database Theory—ICDT 2001: 8th International Conference London, UK, January 4--6, 2001 Proceedings 8}.\hskip 1em plus 0.5em minus 0.4em\relax Springer, 2001, pp. 420--434.

\bibitem{full1999templates}
R.~J. Full and D.~E. Koditschek, ``Templates and anchors: neuromechanical hypotheses of legged locomotion on land,'' \emph{Journal of experimental biology}, vol. 202, no.~23, pp. 3325--3332, 1999.

\bibitem{herreid1986locomotion}
C.~F. Herreid and R.~J. Full, ``Locomotion of hermit crabs (coenobita compressus) on beach and treadmill,'' \emph{Journal of Experimental Biology}, vol. 120, no.~1, pp. 283--296, 1986.

\bibitem{daley2007proxdist}
M.~A. Daley, G.~Felix, and A.~A. Biewener, ``Running stability is enhanced by a proximo-distal gradient in joint neuromechanical control,'' \emph{Journal of Experimental Biology}, vol. 210, no.~3, pp. 383--394, 2007.

\bibitem{berniker2009simplified}
M.~Berniker, A.~Jarc, E.~Bizzi, and M.~C. Tresch, ``Simplified and effective motor control based on muscle synergies to exploit musculoskeletal dynamics,'' \emph{Proceedings of the National Academy of Sciences}, vol. 106, no.~18, pp. 7601--7606, 2009.

\bibitem{valero2016fundamentals}
F.~J. Valero-Cuevas, \emph{Fundamentals of neuromechanics}.\hskip 1em plus 0.5em minus 0.4em\relax Springer, 2016, vol.~8.

\end{thebibliography}

\end{document}